\documentclass{article}

\usepackage{arxiv}

\usepackage[utf8]{inputenc} 
\usepackage[T1]{fontenc}    
\usepackage{hyperref}       
\usepackage{url}            
\usepackage{booktabs}       
\usepackage{amsfonts}       
\usepackage{nicefrac}       
\usepackage{microtype}      
\usepackage{lipsum}		
\usepackage{graphicx}
\usepackage{natbib}
\usepackage{doi}
\usepackage{subcaption}

\title{Image-Scale Robustness and Visual Recognition Performance: A Cross-Architecture Analysis}

\date{} 					

\author{
Anish Monsley Kirupakaran\,
\href{https://orcid.org/0000-0002-4927-3785}
{\includegraphics[height=1.8ex]{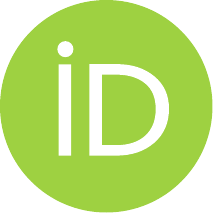}}\\
\texttt{anishmonsley@yahoo.com}
}
    
\renewcommand{\shorttitle}{\textit{Image-Scale Robustness and Visual Recognition Performance}}

\begin{document}
\maketitle

\begin{abstract}
The sensitivity of visual recognition models to changes in image scale is well established, yet the factors governing this sensitivity across heterogeneous architectures remain unclear. In this work, we investigate whether scale robustness exhibits a common quantitative structure across modern vision models. We evaluate 20 pretrained ImageNet-1K classifiers spanning seven architectural families, including convolutional, mobile, efficient, and Transformer-based architectures. By systematically reducing input image scale, we construct scale--accuracy response curves and define a characteristic scale as a compact measure of the onset of substantial recognition degradation. We then examine the relationship between characteristic scale and baseline recognition accuracy, model parameter count, architectural family, and representation stability. A strong inverse association is observed between baseline accuracy and characteristic scale (Pearson $r=-0.890$, $R^2=0.792$, $p<10^{-6}$). This relationship remains stable under bootstrap resampling, leave-one-architecture-out analysis, and leave-one-family-out analysis. In contrast, parameter count provides negligible additional explanatory power after controlling for baseline accuracy ($p=0.80$), while architectural family does not provide significant incremental explanatory power. Furthermore, characteristic scale shows essentially no association with representation stability ($r=-0.003$, $p=0.991$). These results indicate that, across the studied models, scale robustness is strongly organized by baseline recognition performance rather than simply by model size, architectural family, or representation stability. The study provides an empirical framework for characterizing scale robustness across vision architectures and identifies a reproducible accuracy--scale regularity that warrants further theoretical investigation.
\end{abstract}


\section{Introduction}

Scale variation has long been recognized as a fundamental challenge in visual recognition. Classical scale-space methods explicitly represent image structures across multiple spatial scales \citep{Lindeberg}, while subsequent deep-learning approaches have sought to improve scale invariance through multi-scale processing, shared representations, and scale-aware architectures \citep{Aharon_Azulay,Takahashi1,Robert}. In particular, weight-shared multi-stage convolutional networks have demonstrated improved robustness to object scaling on ImageNet \citep{Takahashi1}, and scale-channel architectures have been developed to obtain invariant representations over previously unseen scales \cite{Kanazawa2014}. These studies establish that conventional recognition networks can exhibit substantial sensitivity to scale and that explicit architectural mechanisms can improve scale robustness.

    More recent studies have examined scale and resolution robustness in contemporary vision architectures from several complementary perspectives. Researchers have investigated robustness under distribution shifts while varying training data, model size, and input resolution, showing that changes in image resolution can substantially influence robustness \citep{djolonga2021}. Furthermore, researchers have studied the robustness of Vision Transformers and convolutional networks across several perturbation benchmarks and reported systematic relationships between model size and robustness \citep{bhojanapalli2021understanding}. Other studies have investigated resolution scalability of Vision Transformers through deeper architectures, multi-resolution training, and test-time resolution changes \citep{Touvron_2021_ICCV}, while recent work has examined the mechanisms underlying scale-aware and hierarchical Transformer architectures \citep{liu2021}. These studies provide important evidence that input scale, model capacity, and architectural design influence robustness.

    However, an important question remains largely unexplored. Existing studies primarily evaluate robustness within individual architectures, compare selected architectures, or develop mechanisms for improving scale invariance and resolution generalization. These works do not investigate whether scale robustness can be represented by a common quantitative variable across heterogeneous modern architectures, nor whether variation in such a variable is systematically explained by baseline recognition performance rather than by parameter count or architectural family. In particular, it remains unclear whether the scale at which recognition performance deteriorates should be regarded primarily as an architecture-specific property or as a systematic property associated with the recognition capability of the model.

    This distinction motivates the present study. We introduce a controlled scale-response characterization in which the performance of a pretrained classifier is measured continuously as the spatial scale of its input is reduced. From the resulting response curve, we define a \emph{characteristic scale}, providing a compact measure of the scale tolerance of a recognition model. We then evaluate this quantity across 20 ImageNet-1K pretrained architectures spanning seven distinct architectural families, including ResNet \citep{He}, DenseNet \citep{Huang}, MobileNetV2 \citep{Sandler}, EfficientNet \citep{Tan}, ConvNeXt, Vision Transformer (ViT) \citep{dosovitskiy}, and Swin Transformer \citep{Liu}. Rather than proposing a new architecture or training procedure, our objective is to determine whether characteristic scale exhibits a reproducible cross-architecture relationship and to identify which measurable properties of a model account for its variation.

We specifically examine four potential explanatory factors: baseline recognition accuracy, parameter count, architectural family, and representation stability under scale reduction. This formulation allows scale robustness to be examined as a statistical property of a population of models rather than as a performance characteristic of a single architecture. Through cross-architecture analysis and statistical validation, we investigate whether the characteristic scale is systematically associated with recognition performance and whether model size, architectural family, or representation stability provide additional explanatory power.

\section{Methods and Experimental Framework}

\subsection{Research Question}
This study investigates the determinants of image-scale robustness in pretrained image classification models. In particular, we ask whether a model's tolerance to input-scale reduction can be explained primarily by its baseline recognition capability, its parameter count, or its architectural design. We further examine whether differences in scale robustness are reflected in the stability of the learned representations. Rather than comparing robustness only through absolute accuracy at individual image scales, we characterize each model using a single scale-dependent quantity, termed the characteristic scale. This provides a common basis for comparing architectures with substantially different baseline accuracies and model sizes.

\subsection{Models and Dataset}

Experiments were conducted using a fixed set of 20 pretrained image-classification architectures spanning seven architectural families: ResNet \citep{He}, MobileNet \citep{Sandler, Howard2019}, DenseNet \citep{Huang}, EfficientNet \citep{Tan}, Swin Transformer \citep{Liu}, Vision Transformer (ViT) \citep{dosovitskiy}, and ConvNeXt \citep{ConvNeXt}. The evaluated models were ResNet18, ResNet34, ResNet50, ResNet101, ResNet152, MobileNetV2, MobileNetV3-Large, DenseNet121, DenseNet169, DenseNet201, EfficientNet-B0, EfficientNet-B1, EfficientNet-B2, EfficientNet-B3, EfficientNet-B4, Swin-T, ViT-B/16, ConvNeXt-Tiny, ConvNeXt-Small, and ConvNeXt-Base. All models were evaluated using the same ImageNet-1K validation data and the same image preprocessing and evaluation protocol. The models span approximately 3.5–88.6 million trainable parameters, providing substantial variation in both model capacity and architectural structure. Baseline Top-1 accuracy was measured at the reference image scale before applying scale perturbations.

\subsection{Image-Scale Perturbation and Characteristic Scale}
For each architecture, the input images were evaluated over a range of systematically reduced spatial scales. Let $s \in (0,1]$ denote the relative image scale, where $s=1$ represents the reference input resolution. Let $A_i(s)$ denote the Top-1 accuracy of model $i$ at scale $s$, and let $A_{i,\mathrm{base}}$ denote its baseline accuracy at the reference scale.

To provide a model-independent measure of scale tolerance, we define the
normalized scale--response function as
\begin{equation}
R_i(s)
=
\frac{A_i(s)}{A_{i,\mathrm{base}}}.
\label{eq:normalized_scale_response}
\end{equation}

The characteristic scale, $S_{c,i}$, is defined as the smallest relative
scale at which the normalized performance satisfies a predefined retention
criterion $\tau$:
\begin{equation}
S_{c,i}
=
\min
\left\{
s \in (0,1] :
R_i(s) \geq \tau
\right\}.
\label{eq:characteristic_scale}
\end{equation}

Thus, $S_{c,i}$ represents the smallest relative image scale at which model
$i$ retains the specified fraction $\tau$ of its baseline recognition
performance. A larger $S_{c,i}$ indicates that a model requires a larger
input representation to maintain its performance, whereas a smaller
$S_{c,i}$ indicates greater tolerance to scale reduction.

The same scale--response procedure was applied uniformly across all 20
architectures, allowing characteristic scales to be compared independently
of differences in model implementation.

\subsection{Determining the Factors Associated with Scale Robustness}
We first examined the relationship between characteristic scale and baseline
accuracy using Pearson and Spearman correlation analyses. Ordinary
least-squares (OLS) regression was then used to quantify the association:
\begin{equation}
S_c = \beta_0 + \beta_1 A_{\mathrm{base}} + \epsilon,
\label{eq:accuracy_scale_regression}
\end{equation}
where $S_c$ denotes the characteristic scale, $A_{\mathrm{base}}$ denotes
baseline Top-1 accuracy, and $\epsilon$ represents the residual error.

To determine whether model capacity provides additional explanatory power, a
second model incorporated the logarithm of the number of trainable parameters:
\begin{equation}
S_c = \beta_0 + \beta_1 A_{\mathrm{base}}
      + \beta_2 \log(P) + \epsilon,
\label{eq:accuracy_parameter_regression}
\end{equation}
where $P$ denotes the number of trainable model parameters. The incremental
contribution of model size was evaluated by comparing this model with the
accuracy-only model.

An extended model additionally incorporated architecture family as a
categorical variable:
\begin{equation}
S_c = \beta_0 + \beta_1 A_{\mathrm{base}}
      + \sum_{k=1}^{K-1}\gamma_k F_k + \epsilon,
\label{eq:architecture_family_regression}
\end{equation}
where $F_k$ denotes the indicator variable for the $k$th architecture family,
with one family treated as the reference category. The incremental
contribution of architecture family was evaluated using a partial
$F$-test and permutation testing.

\subsection{Accuracy-Matched and Robustness-Residual Analysis}
Because baseline accuracy and scale robustness may be intrinsically related, we further performed accuracy-matched comparisons between architectures. Models with similar baseline accuracies were paired using predefined accuracy tolerances, and their characteristic-scale differences were examined.

We additionally quantified architecture-specific deviations from the accuracy-based expectation. For each model, the robustness residual was defined as
\begin{equation}
R_i = S_{c,i} - \widehat{S}_{c,i},
\label{eq:robustness_residual}
\end{equation}
where $S_{c,i}$ is the observed characteristic scale of model $i$ and $\widehat{S}_{c,i}$ is the characteristic scale predicted from the baseline-accuracy relationship. Positive residuals therefore indicate models that exhibit greater scale tolerance than expected from their baseline accuracy, whereas negative residuals indicate lower-than-expected scale tolerance.

This analysis separates the dominant accuracy-associated component of scale robustness from architecture-specific deviations, allowing models with comparable baseline performance to be evaluated in terms of their relative scale robustness.

\subsection{Statistical Robustness Analysis}
The stability of the observed accuracy–scale relationship was assessed using nonparametric bootstrap resampling with 10,000 replicates. Confidence intervals were obtained for Pearson correlation, Spearman correlation, regression coefficients, and $R^2$. Permutation tests with 5,000 permutations were used to evaluate whether the observed associations could arise under a null relationship. In addition, leave-one-architecture-out (LOO) and leave-one-family-out analyses were performed to determine whether the principal relationship was driven disproportionately by any individual architecture or architectural family.

\subsection{Representation Stability Analysis}
We also investigated whether scale robustness could be explained by the stability of the internal representations produced by the models. For each architecture, representation stability across image scales was quantified from the extracted feature representations and summarized using a representation-stability area-under-the-curve (AUC) measure. The resulting representation-stability measure was compared against characteristic scale using Pearson and Spearman correlation, permutation testing, bootstrap confidence intervals, and leave-one-architecture/family-out analyses. This final analysis tests whether models that maintain more stable internal representations under scale reduction necessarily exhibit greater image-scale robustness.

\section{Results and Discussion}
\subsection{Characteristic Scale is Strongly Associated with Baseline Recognition Accuracy}

We first examined whether a model's tolerance to image-scale reduction is related to its baseline recognition capability. Across the 20 evaluated architectures, characteristic scale exhibited a strong negative association with baseline Top-1 accuracy. Pearson correlation yielded $r=-0.890$ ($p=1.52\times10^{-7}$), while the corresponding Spearman rank correlation was $\rho=-0.882$ ($p=2.82\times10^{-7}$). The resulting regression explained approximately 79.2\% of the variance in characteristic scale ($R^2=0.792$).

The fitted relationship was

\begin{equation}
S_c = 1.639 - 1.282A_{\mathrm{base}}.
\end{equation}

Thus, within the evaluated model set, architectures with higher baseline recognition accuracy generally reached the characteristic-scale criterion at smaller image scales. Importantly, this relationship is not attributable to a small number of observations. A permutation test with 5,000 permutations yielded $p=2.0\times10^{-4}$, providing strong evidence against the null hypothesis of no accuracy--scale association.

\begin{figure}[t]
    \centering
    \includegraphics[width=0.85\linewidth]{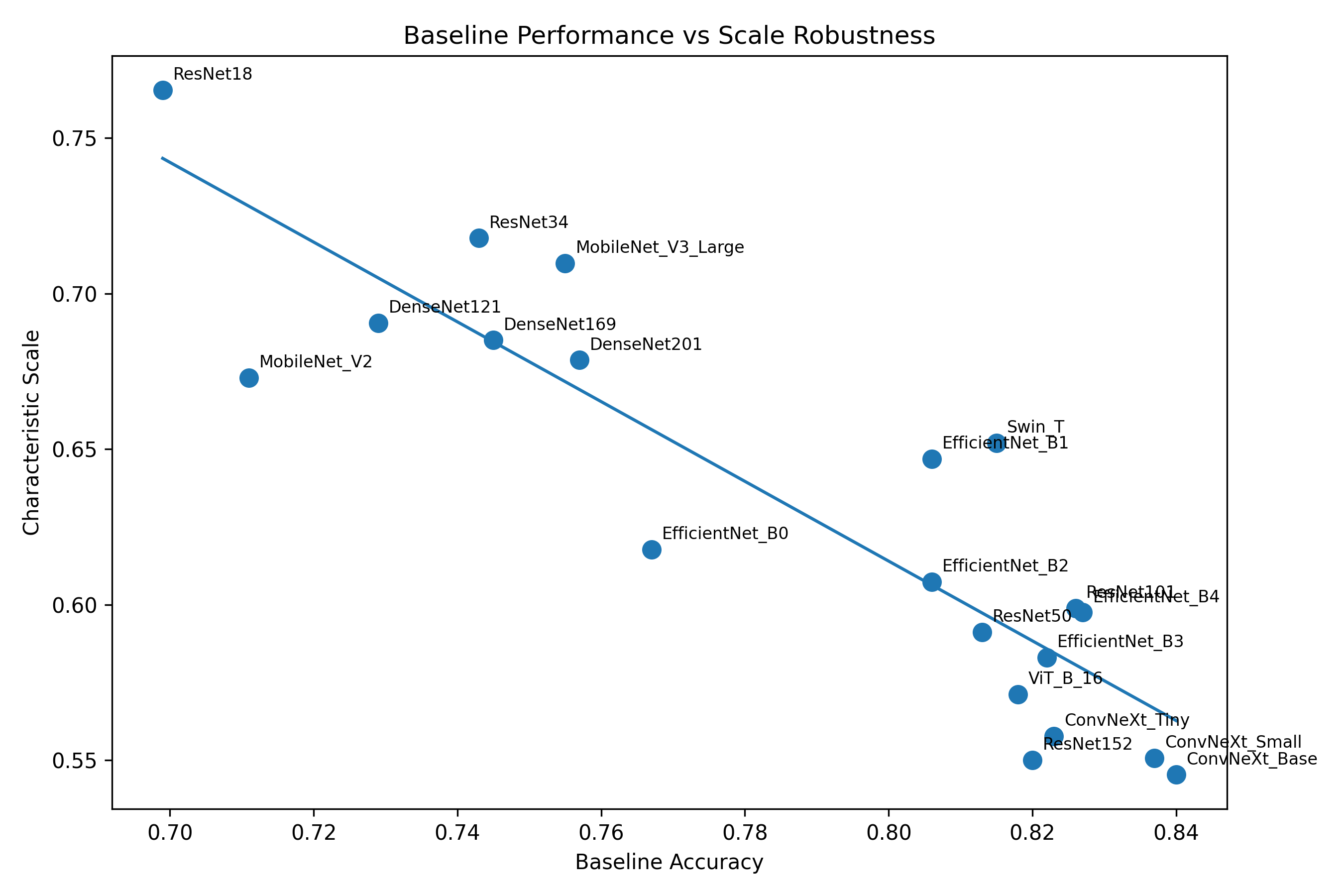}
    \caption{Each point represents one architecture. The fitted regression demonstrates a strong negative association between baseline accuracy and characteristic scale ($r=-0.890$, $R^2=0.792$, $p<10^{-6}$). Higher-performing architectures generally retain their recognition performance at smaller image scales.}
    \label{fig:Relationship between baseline recognition accuracy and characteristic image scale across 20 pretrained architectures}
\end{figure}

The uncertainty analysis further supports the stability of this observation. Bootstrap resampling with 10,000 valid samples produced a 95\% confidence interval of $[-0.961,\,-0.776]$ for Pearson's $r$, with the correlation remaining negative in all bootstrap samples. The corresponding interval for Spearman's $\rho$ was $[-0.956,\,-0.695]$, while the bootstrap interval for $R^2$ was $[0.602,\,0.924]$. These results indicate that the observed association is strong and statistically stable despite the relatively small number of architectures.

\subsection{Model Capacity Provides Little Additional Explanatory Power}
We next investigated whether the observed relationship could instead be attributed primarily to model size. When characteristic scale was regressed only on the logarithm of parameter count, parameter count explained 37.3\% of the observed variance ($R^2=0.373$), with a significant negative coefficient. However, introducing baseline accuracy into the model substantially changed this interpretation. The combined model,

\begin{equation}
S_c = 1.619 - 1.246A_{\mathrm{base}} - 0.006\log(P),
\end{equation}

achieved $R^2=0.793$. Adding parameter count to the accuracy-only model increased $R^2$ by only approximately 0.001. More importantly, the parameter coefficient was not statistically significant ($p=0.80$). Thus, after controlling for baseline accuracy, model size provided essentially no additional explanatory power for characteristic scale. This distinction is important because parameter count is correlated with several aspects of modern architecture design. The results suggest that scale robustness cannot simply be interpreted as a consequence of increasing model capacity. Within the evaluated models, baseline recognition performance provides substantially more explanatory information about characteristic scale than parameter count.

\subsection{The Accuracy–Scale Relationship is Not Driven by Individual Architectures or Families}

\begin{figure}[t]
    \centering

    \begin{subfigure}[b]{0.48\linewidth}
        \centering
        \includegraphics[width=\linewidth]{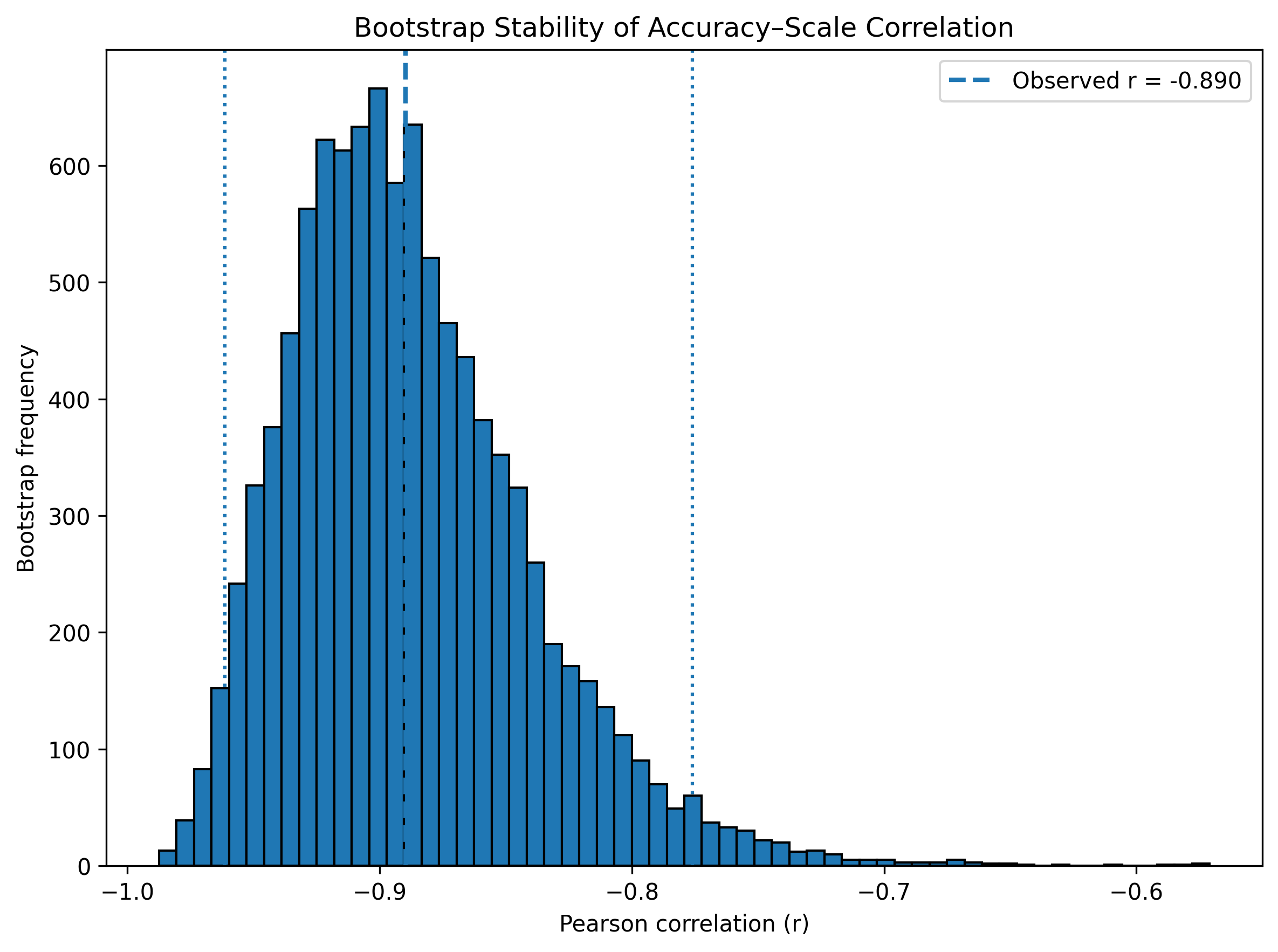}
        \caption{Bootstrap distribution of Pearson correlation.}
        \label{fig:bootstrap_accuracy_scale}
    \end{subfigure}
    \hfill
    \begin{subfigure}[b]{0.48\linewidth}
        \centering
        \includegraphics[width=\linewidth]{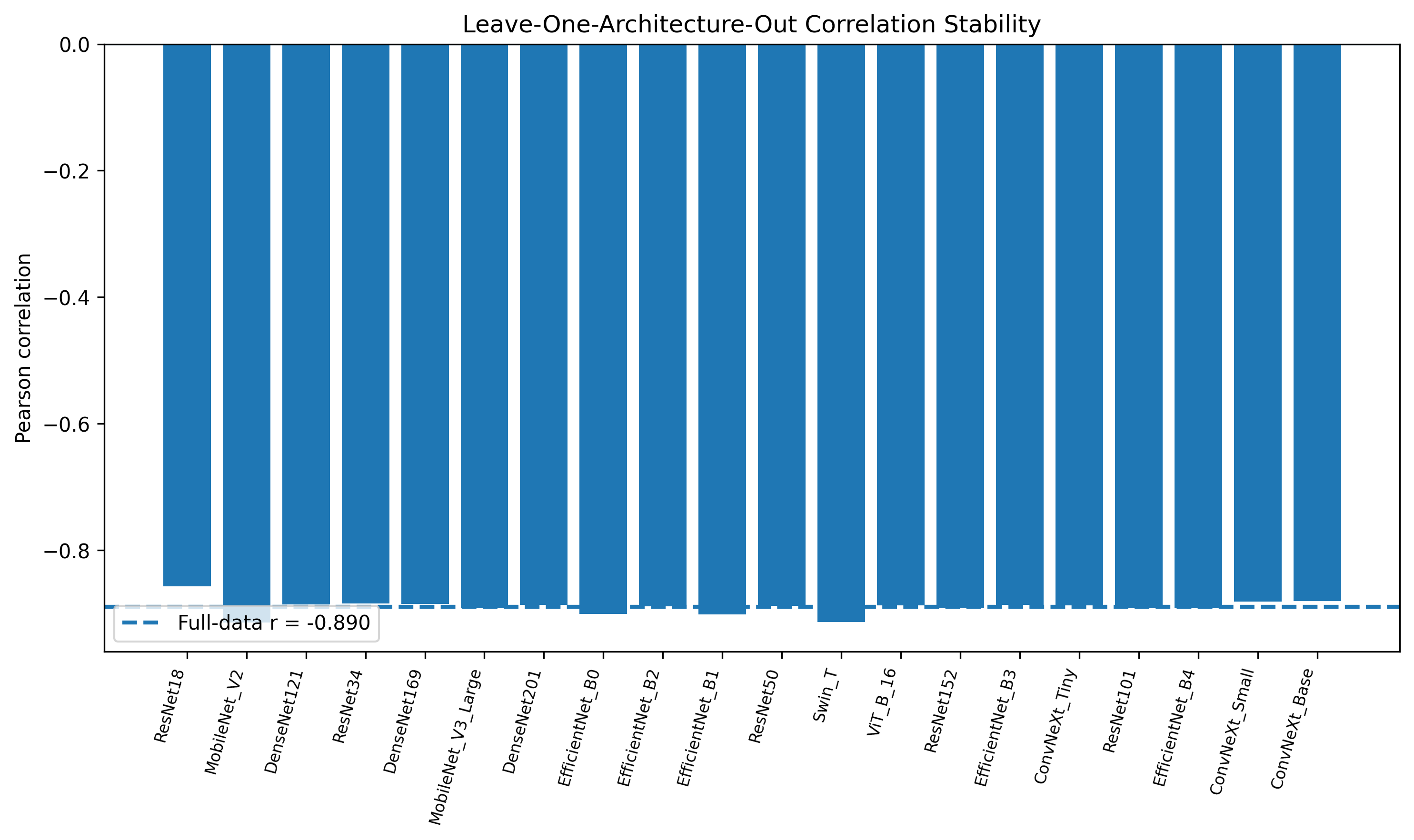}
        \caption{Leave-one-architecture-out correlations.}
        \label{fig:loo_architecture}
    \end{subfigure}

    \caption{Robustness of the accuracy--scale relationship. (a) Bootstrap distribution of Pearson correlation between baseline accuracy and characteristic scale over 10,000 resamples, with the observed correlation and 95\% confidence interval indicated. (b) Leave-one-architecture-out correlations demonstrate that the negative association persists across all individual architecture exclusions.}
    \label{fig:robustness_accuracy_scale}
\end{figure}

The robustness of the accuracy--scale relationship was further evaluated using bootstrap resampling and leave-one-architecture-out analysis (Fig.~\ref{fig:robustness_accuracy_scale}). To test whether the observed relationship was dominated by particular architectures, we performed a leave-one-architecture-out analysis. Removing any individual model produced Pearson correlations ranging from $-0.914 \leq r \leq -0.857$. The corresponding $R^2$ values remained substantial across all exclusions. The weakest association occurred when ResNet18 was removed ($r=-0.857$, $R^2=0.734$), while the strongest occurred when MobileNetV2 was removed ($r=-0.914$, $R^2=0.836$).

We further performed a leave-one-family-out analysis. The Pearson correlation remained negative for every family exclusion, ranging from $-0.915$ to $-0.845$. Therefore, the principal accuracy--scale relationship persists even when an entire architectural family is excluded. We examined whether architecture family itself provides an independent explanation for characteristic scale. An extended model incorporating architecture family increased the nominal $R^2$ from $0.792$ to $0.859$. However, this improvement was not statistically significant: the partial $F$-test gave $F=0.863$ with $p=0.550$, while a 5,000-permutation test produced $p=0.562$. Consequently, the apparent increase in explained variance from adding architectural family should not be interpreted as evidence that family identity is a statistically established determinant of characteristic scale. Rather, the results indicate that architecture-specific differences exist, but the dominant cross-architecture trend is already captured by baseline accuracy.

\subsection{Architecture-Specific Deviations Reveal Models That Exceed Accuracy-Based Expectations}
\begin{figure}[t]
    \centering
    \includegraphics[width=0.85\linewidth]{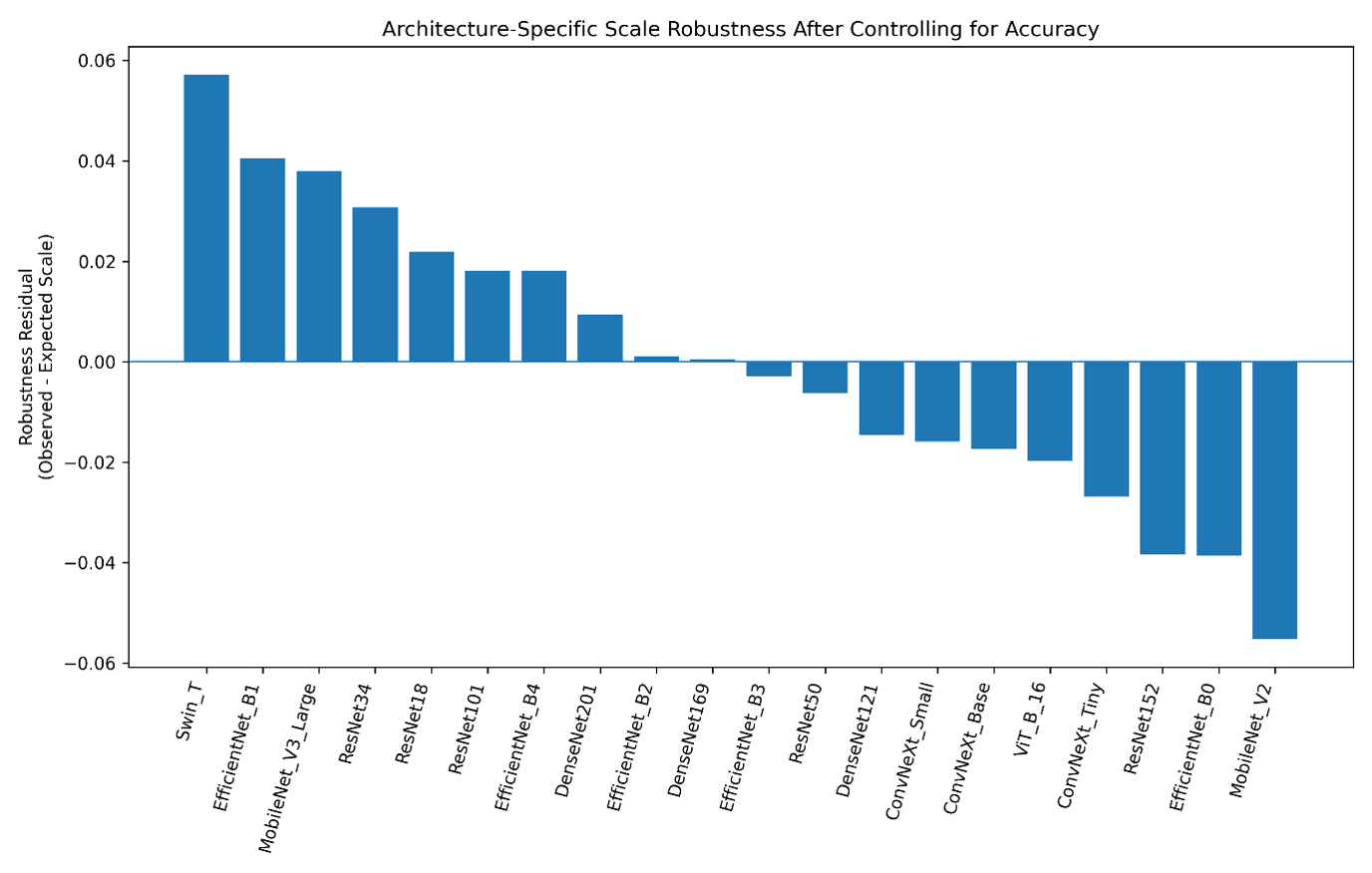}
    \caption{Robustness residuals quantify the difference between observed characteristic scale and the value predicted from baseline accuracy. Positive residuals indicate greater-than-expected scale tolerance, whereas negative residuals indicate lower-than-expected tolerance..}
    \label{fig:Architecture-specific deviations from the accuracy-based robustness relationship}
\end{figure}

Although baseline accuracy explains much of the variation in characteristic scale, it does not explain all of it. We therefore examined the residual robustness of individual architectures relative to the accuracy-based prediction (refer Fig.~\ref{fig:Architecture-specific deviations from the accuracy-based robustness relationship}).

The largest positive residuals were observed for Swin-T (+0.057), EfficientNet-B1 (+0.040), MobileNetV3-Large (+0.038), ResNet34 (+0.031), and ResNet18 (+0.022). These architectures exhibited greater scale tolerance than would be expected solely from their baseline accuracy. In contrast, MobileNetV2 (-0.055), EfficientNet-B0 (-0.039), ResNet152 (-0.038), ConvNeXt-Tiny (-0.027), and ViT-B/16 (-0.020) exhibited lower characteristic scales than predicted by the accuracy-only relationship. Notably, the residual robustness showed essentially no relationship with parameter count (Pearson r=-0.046, p=0.846). This provides additional evidence that the remaining architecture-specific variation cannot simply be attributed to model size. These residuals therefore provide a useful distinction between two effects: (1) a dominant performance-associated component of scale robustness; (2) a smaller architecture-specific component that determines whether an individual model performs above or below that expectation.

\subsection{Representation Stability Does Not Explain Characteristic Scale}

We also investigated whether the observed scale robustness could be explained by stability of the internal feature representations. If representation preservation were the primary mechanism underlying scale robustness, models with higher representation-stability AUC would be expected to exhibit systematically larger characteristic scales.

\begin{figure}[t]
    \centering
    \includegraphics[width=0.85\linewidth]{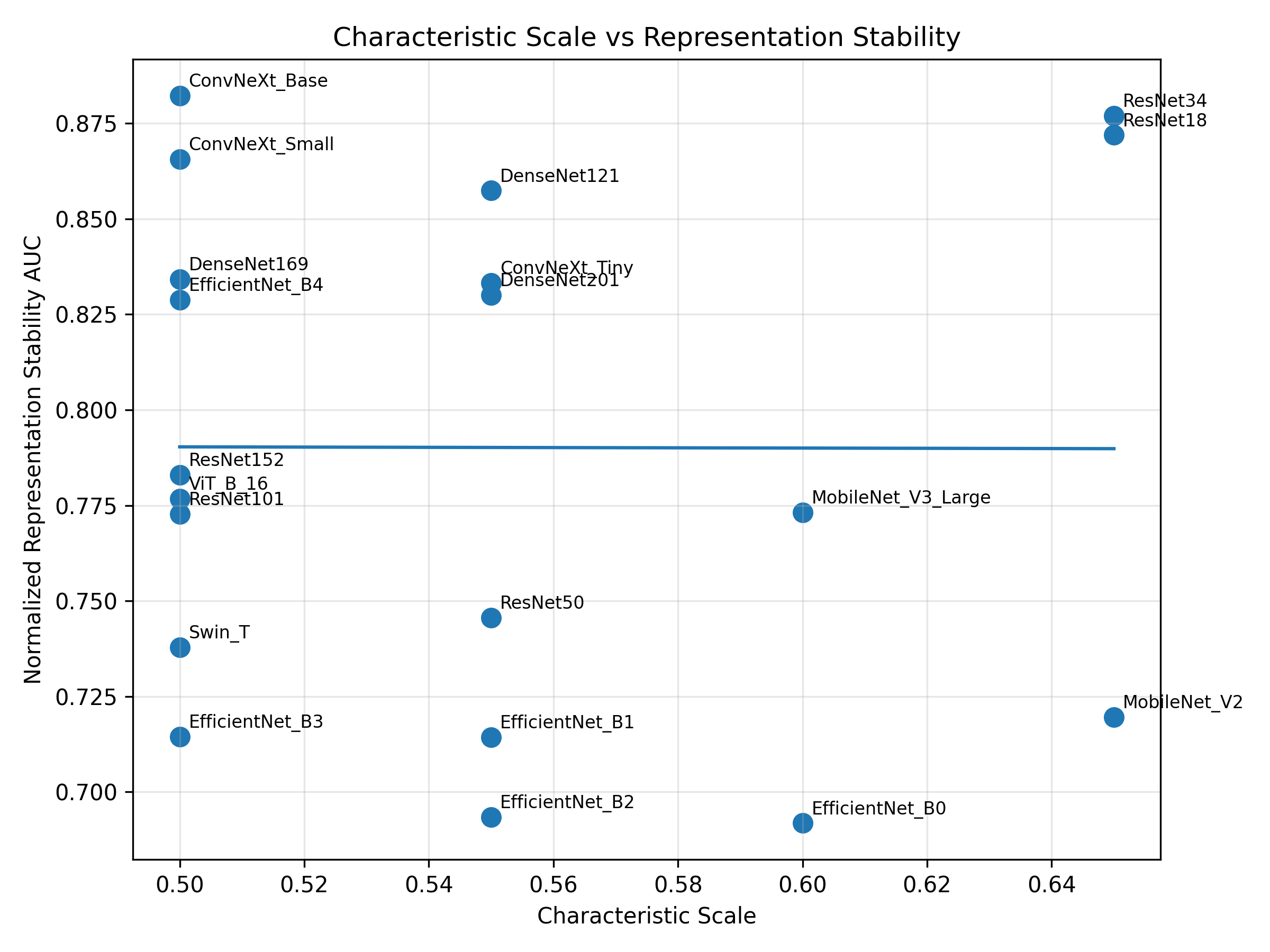}
    \caption{Characteristic scale versus representation stability. No significant association is observed between characteristic scale and representation-stability AUC ($r=-0.003$, $p=0.991$), indicating that the observed cross-architecture variation in scale robustness is not explained by representation stability alone.}
    \label{fig:scale_representation_stability}
\end{figure}

As shown in Fig.~\ref{fig:scale_representation_stability}, no significant association is observed between characteristic scale and representation-stability AUC. Across the 20 architectures, Pearson correlation between characteristic scale and representation-stability AUC was $r=-0.003$, $p=0.991$, while Spearman correlation was $\rho=-0.063$, $p=0.791$. A 5,000-permutation test yielded $p=0.497$. Bootstrap analysis with 10,000 samples produced a 95\% confidence interval of $[-0.528,\,0.480]$ for Pearson's $r$, which includes both moderate positive and negative associations. The probability of obtaining a negative correlation across bootstrap samples was approximately 0.51. The absence of association was also robust to model and family exclusions. Leave-one-architecture-out Pearson correlations ranged from $-0.164$ to $0.126$, while leave-one-family-out correlations ranged from $-0.434$ to $0.154$. These results indicate that representation stability alone is not a sufficient explanation for the observed characteristic-scale differences. In particular, the strong accuracy--scale relationship established above does not appear to arise simply because higher-performing models preserve more stable feature representations under scale reduction.

\subsection{Findings}
These experiments reveal three principal findings. 
\begin{itemize}
\item Characteristic scale exhibits a strong and statistically robust relationship with baseline recognition accuracy, explaining approximately 79\% of its observed variation across the evaluated architectures. 
\item This relationship remains stable under architecture- and family-level exclusion, while parameter count contributes negligible additional explanatory power after accounting for baseline accuracy. 
\item  Representation stability does not exhibit a measurable cross-architecture association with characteristic scale.
\end{itemize}
The results therefore suggest that image-scale robustness is neither a simple consequence of parameter count nor directly determined by representation stability. Instead, the dominant empirical predictor observed in this study is the model's baseline recognition capability, with architecture-specific residuals accounting for additional variation not captured by accuracy alone.

\section{Limitations}

\begin{itemize}
\item Characteristic scale is an empirical descriptor derived from the scale–response curve; it should not be interpreted as a universal architectural constant.
\item The strong relationship between baseline accuracy and characteristic scale establishes a robust statistical association, but does not by itself identify the underlying causal mechanism.
\item The absence of a relationship with the aggregate representation-stability measure does not rule out more localized or layer-specific mechanisms contributing to scale robustness.
\item The consistency of the relationship across architectures and multiple statistical robustness tests supports its reliability within the present study, while broader validation across datasets and visual tasks remains an important direction for future work.
\end{itemize}

\section{Conclusion}
This study investigated how image-scale robustness varies across modern visual recognition architectures and, more importantly, whether this variation can be explained by model recognition capability, capacity, or architectural family. Across 20 pretrained architectures, we observed a strong and consistent inverse relationship between baseline recognition accuracy and characteristic scale, with higher-performing models generally retaining reliable recognition at smaller image scales. The relationship remained stable under bootstrap analysis, permutation testing, and systematic leave-one-architecture-out and leave-one-family-out analyses, indicating that the observed trend is not dominated by individual models or a particular architectural family. In contrast, parameter count provided substantially weaker explanatory power, and its contribution largely disappeared after accounting for baseline accuracy. Architectural family also provided limited additional explanatory power once recognition accuracy was considered. These results suggest that scale robustness is more closely associated with recognition capability than with model size alone or architectural taxonomy. The resulting characteristic-scale formulation provides a simple quantitative framework for comparing scale robustness across heterogeneous vision models. More broadly, the findings indicate that image-scale tolerance may represent an emergent property associated with recognition performance rather than a direct consequence of parameter count or architecture family. This perspective opens a pathway toward systematic study of scale robustness as an independent property of visual recognition systems.

\section*{Declarations}
\subsection*{Funding}
The author received no specific funding for this work.

\subsection*{Conflict of Interest }
The author declare that they have no known competing financial interests or personal relationships that could have appeared to influence the work reported in this manuscript.

\subsection*{Ethics Approval}
Not applicable. 


\subsection*{Data Availability}
The datasets analyzed during the current study are publicly available. The ImageNet-1K dataset was used in accordance with its licensing and access policies. Processed data and additional results supporting the findings of this study are available from the corresponding author upon reasonable request.

\subsection*{Reproducibility Statement}
To support the reproducibility of the reported findings, the source code, evaluation scripts, statistical analysis scripts, and figure-generation pipelines used in this study can be shared by the corresponding author upon reasonable request. These materials are sufficient to reproduce the experiments, analyses, and results reported in the manuscript.

\bibliographystyle{unsrtnat}
\bibliography{references}  






\end{document}